\documentclass[letterpaper, 10 pt, conference]{ieeeconf}
\usepackage{subcaption}
\usepackage{amsmath} % assumes amsmath package installed
\usepackage{amssymb}  % assumes amsmath package installed
\usepackage{mathtools}
\usepackage{graphicx}
\usepackage{float}
\usepackage[hyphens]{url}
\usepackage[hidelinks]{hyperref}
\usepackage{booktabs}
\usepackage{csvsimple}
\usepackage{longtable}
\usepackage{multirow}
\usepackage{xcolor}
\usepackage{wrapfig}
\usepackage{multicol}

\usepackage{array}    % from https://tex.stackexchange.com/a/616713
\usepackage{varwidth} % must be loaded after array

\usepackage[]{pgfplots}

\usepgfplotslibrary{fillbetween}
\pgfplotsset{
    compat=newest, 
    every axis/.append style={
        grid=both,
        major grid style={lightgray},
        minor grid style={lightgray!15}
    },
    minor tick num=3,
    every minor tick/.style={minor tick length=0pt}
}
\usetikzlibrary{calc}

\hypersetup{breaklinks=true}
\usepackage{cleveref}
\usepackage{algorithm}
\usepackage{fancyvrb}

\newcommand{\R}{\mathbb{R}}
\newcommand{\sethree}{\ensuremath{\text{SE}(3)}}

\usepackage{xcolor}

\IEEEoverridecommandlockouts
\title{\LARGE \bf 
CppFlow: Generative Inverse Kinematics for Efficient and Robust Cartesian Path Planning
}

\author{Jeremy Morgan* \and David Millard* \and Gaurav S. Sukhatme
  \thanks{*Equal contribution. All authors are with the Department of Computer Science at the University of Southern California. GSS holds concurrent appointments as a Professor at USC and as an Amazon Scholar. This paper describes work performed at USC and is not associated with Amazon. This work is supported in part by the NASA Space Technology Research Fellowship, grant number 80NSSC19K1182.}
}
\begin{document}
\maketitle % needs to come after \author{}

% ICRA page limit: "The page limit is 6 pages for the paper (text, figures, tables, acknowledgement, etc.) + any number of pages for the bibliography/references" (https://2024.ieee-icra.org/call-for-contributions.html)

% Update planning time table and convergence plots with:
% python3 scripts/save_convergence_data.py
% python3 scripts/save_planning_time_table.py

%===============================================================================

\begin{abstract}
	In this work we present CppFlow - a novel and performant planner for the Cartesian Path Planning problem, which finds valid trajectories up to 129x faster than current methods, while also succeeding on more difficult problems where others fail. At the core of the proposed algorithm is the use of a learned, generative Inverse Kinematics solver, which is able to efficiently produce promising entire candidate solution trajectories on the GPU. Precise, valid solutions are then found through classical approaches such as differentiable programming, global search, and optimization. In combining approaches from these two paradigms we get the best of both worlds - efficient approximate solutions from generative AI which are made exact using the guarantees of traditional planning and optimization. We evaluate our system against other state of the art methods on a set of established baselines as well as new ones introduced in this work and find that our method significantly outperforms others in terms of the time to find a valid solution and planning success rate, and performs comparably in terms of trajectory length over time. Additional results and an open source implementation is available at \href{https://jstmn.github.io/cppflow-website/}{\small \color{blue}{\texttt{https://jstmn.github.io/cppflow-website/}}}.

% \url{https://jstmn.github.io/cppflow-website/}.

% Additionally, we introduce a new set of benchmarks, which are able to elucidate how well
%  a given cartesian path planning algorithm performs on more difficult problems than those commonly used in the literature currently. Through the introduced bias, as well as an efficient search algorithm and trajectory optimizer, our proposed method establishes a new state of the art on this topic by reducing planning time all the while improving robustness.

% Thus we get the best of both worlds, efficient approximate solutions from generative AI which are made exact using the gaurantees of traditional planning and optimization.

% Trajectory optimization methods well but requires a good initial solution. We use generative AI to provide that initial guess. we use differentiable programming, global search, and other optimization routines to refine the suggestions from the precicely valid solutions. 

% On top of this core functionality - several other algorithmic layers are presented, including an optimal search algorithm for finding plans when the algorithm must go down one ‘fork’ in the road of configuration space, and a trajectory optimizer for improving approximate plans.

\end{abstract}

\section{Introduction}

Moving a robot's manipulator along a specified cartesian space path is a fundamental operation in robotics, with applications across nearly all domains. Tasks such as performing a weld, painting a surface, or turning a door handle are all naturally expressed through the declaration of a reference path for the end effector to follow. Further, in settings such as a kitchen, hospital, or assembly line, it is of great importance that a robot be able to \textit{quickly} generate motion plans so as to avoid down time and lost productivity. Thus, the ability to quickly generate smooth, collision free paths for these tasks and in general along provided paths is of great utility, and while a core problem, there are further gains to be made.

% In never before seen environments, its important that 
% The rapid generation of trajectories is of great value in industrial applications when 
%  especially important in many scenarios

% Describe the problem more concretely
More concretely, the Cartesian Path Planning (CPP) problem, otherwise known as Pathwise-Inverse Kinematics (Pathwise-IK) problem is defined whereby the robot must generate smooth, collision free trajectories (including robot-robot and robot-environment collisions) that result in the end effector tracking a specified cartesian space path. In this paper, we consider this task for redundant robots - those with 7 or greater degrees of freedom (DoFs) - which may have an infinite number of IK solutions for a given pose. It is this redundancy which makes the problem difficult, as IK solutions no longer have a discrete family they can be checked against - such as elbow up or down.

% \textcolor{red}{@david can update this sentence to be more precise. Or use this: 'It is this redundancy which makes the problem difficult, as the --- procedure which is possible for non-redundant manipulators can no longer be performed'}

% Thus, the problem is to find a path through configuration space that satisfies these constraints. A distinction must be made between real time planners, and those that plan ahead of time. This work is on the latter. Paths are generated before any motion is made, enabling paths to be verified by safety systems or displayed to human operators.

% Previous approaches
Current state of the art (SOTA) approaches which run in realtime generate motion by formulating and solving a weighted optimization problem~\cite{rakita_relaxedik_2018,9616379,9561505,Wang2023RangedIKAO}. Those that plan trajectories ahead of time either build a graph or perform gradient based optimization~\cite{rakita_stampede_2019,7759814,kang2020torm}. However, while performant for problems where there is a clear basin of good solutions, these methods may get stuck in local optima and fail to find a solution in a reasonable amount of time on more difficult problems.

\begin{figure}[t]
    \centering
    \includegraphics[width=0.36\textwidth]{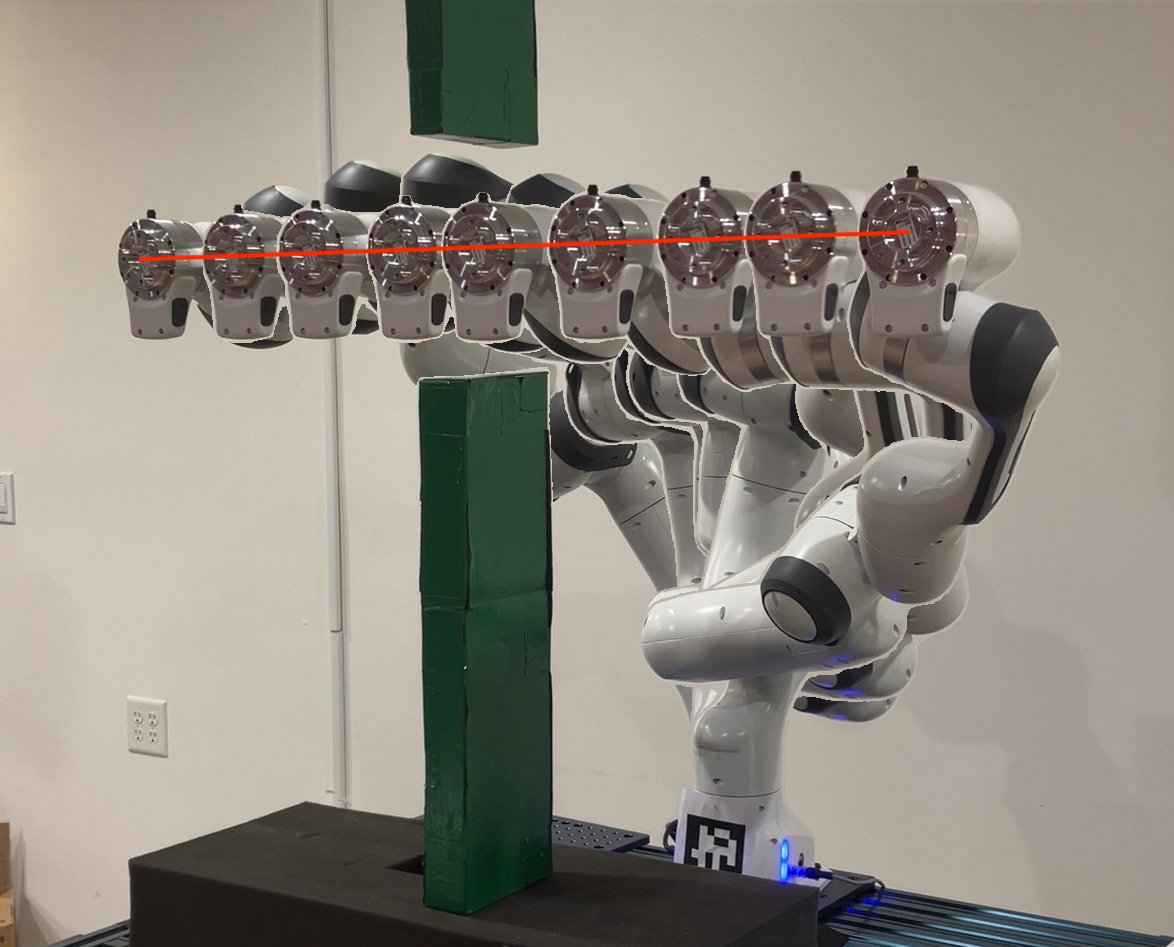}
    \caption{A CppFlow generated trajectory for the `Panda - flappy-bird' problem. In this problem the robot must navigate through a tight corridor imposed by two vertical obstacles.}
    \label{fig:cover_photo}
\end{figure}

% for problems which require global planning. In other terms, purely optimization based methods are likely to fail when there is a proverbial fork in the road in configuration space. Further, there is room for efficiency gains given t
% generate approximate motions by running an interative inverse kinematics (IK) solver before refining trajectories through gradient based optimization. 

\begin{figure*}
    \centering
    \vspace{0.15cm}
    \includegraphics[width=\textwidth]{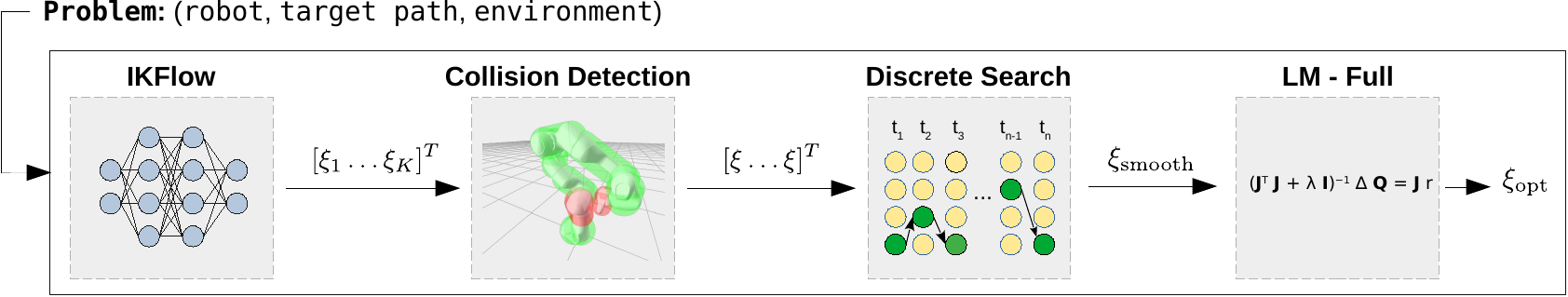}
    \caption{The CppFlow system}
    % When planning a trajectory for a given problem, Cppflow runs three seperate planners in parrallel: \textit{cpf-BatchOpt}, \textit{cpf-Search+FastOpt}, and \textit{cpf-Search+FullOpt} and returns the first valid solution found. Each of the three planners run IKFlow to generate candidate trajectories. \textit{cpf-BatchOpt} optimizes the smoothest 5 of the candidate trajectories directly whereas \textit{cpf-BatchOpt} and \textit{cpf-Search+FastOpt} first perform a global search over all returned configurations to find the optimally smooth path before optimizing. In order, the planners require more planning time but are also more capable. Thus, by running all three at once, Cppflow is able to solve easy problems quickly while still efficiently solving more difficult problems.
    \label[fig]{system_diagram}
    \vspace{-0.7\baselineskip}
\end{figure*}

% Main contribution
In this work we present CppFlow, a novel Cartesian Path Planning planner that utilizes recent advances in learned, generative IK to generate smooth, collision free paths faster than existing SOTA methods while also succeeding on more difficult problems. Using a generative IK model addresses a key issue with trajectory optimization methods - that they work well but require a good initial solution. In this work we use the Levenberg-Marquardt algorithm for trajectory optimization, a powerful quasi-Newton optimization procedure that quickly converges to precise and constraint satisfying solutions. Additionally, a search module finds the optimally smooth config-space path by interweaving the trajectories returned by the generative IK model, which dramatically improves the quality of the optimization seed.

To evaluate CppFlow, we benchmark our method on 5 standard test problems present in the literature. To demonstrate our method's capability on more difficult problems, we introduce a suite of new robots and target paths. The additional robots include the Franka Panda arm and a modified kinematic chain from the Fetch robot that includes a prismatic joint. As opposed to many of the existing tests which contain a basin of valid solutions, the newly introduced tests contain problems that require choosing between disjoint paths in configuration space to reveal the capability of a planner to avoid local minima. Three key metrics of planner performance are reported on: success rate, time to get an initial valid solution, and trajectory length over time. We find that CppFlow outperforms all other planners on two axes and performs comparably on the third. 

% With these problems, Of the three three main axes to grade a cartesian path planner on: 
% For target paths, 

% (added) \textcolor{orange}{Add this in}: Trajectory optimization methods well but requires a good initial solution. We use generative AI to provide that initial guess. we use differentiable programming, global search, and other optimization routines to refine the suggestions from the precicely valid solutions. Thus we get the best of both worlds, efficient approximate solutions from generative AI which are made exact using the gaurantees of traditional planning and optimization.
\section{Problem Specification}

The goal of the Cartesian Path Planning problem is to find a trajectory $\xi$ such that the robots end effector stays along a target path $\boldsymbol{Y}$ while satisfying additional constraints.

It is assumed that the robot has $\ge$7 Degrees of Freedom $d$ and is therefore a redundant manipulator. Owing to this additional freedom, there may be a potentially infinite set of IK solutions for a given end-effector pose. We denote the Forward Kinematics mapping as $\text{FK}(q) : \R^d \rightarrow \sethree$. The target path and trajectory are discretized into $n$ timesteps, such that $\xi = [q_1, ..., q_n]$ and $\boldsymbol{Y} = [y_1, ..., y_n]$, where $q_i \in \R^d$ is the robots joint configuration at timestep $i$ and $y_i \in \sethree$ is the target pose at timestep $i$.

There are two paradigms to addressing time in the Cartesian Path Planning problem. Either a time parameterization is provided with the target path and the goal is to find a satisfying path that minimizes jerk or some other smoothness measure, or no time parameterization is provided, and it is assumed one is found \textit{after} the joint configurations are calculated by some external module. This system uses the latter paradigm - it is assumed that an external module will calculate the temporal component of each $q_i$ such that the robots velocity, acceleration, and jerk limits are respected while execution time is minimized. As such, time parameterization is such omitted from this work. Given this setup, the constraints on the problem are as follows:
% - such that the robot may achieve them in a time optimal manner subject to its velocity, accelleration, and jerk limits

\begin{figure*}
    \centering
    \vspace{0.15cm}
    \begin{subfigure}[b]{0.23\textwidth}
        \centering
        \includegraphics[width=\textwidth]{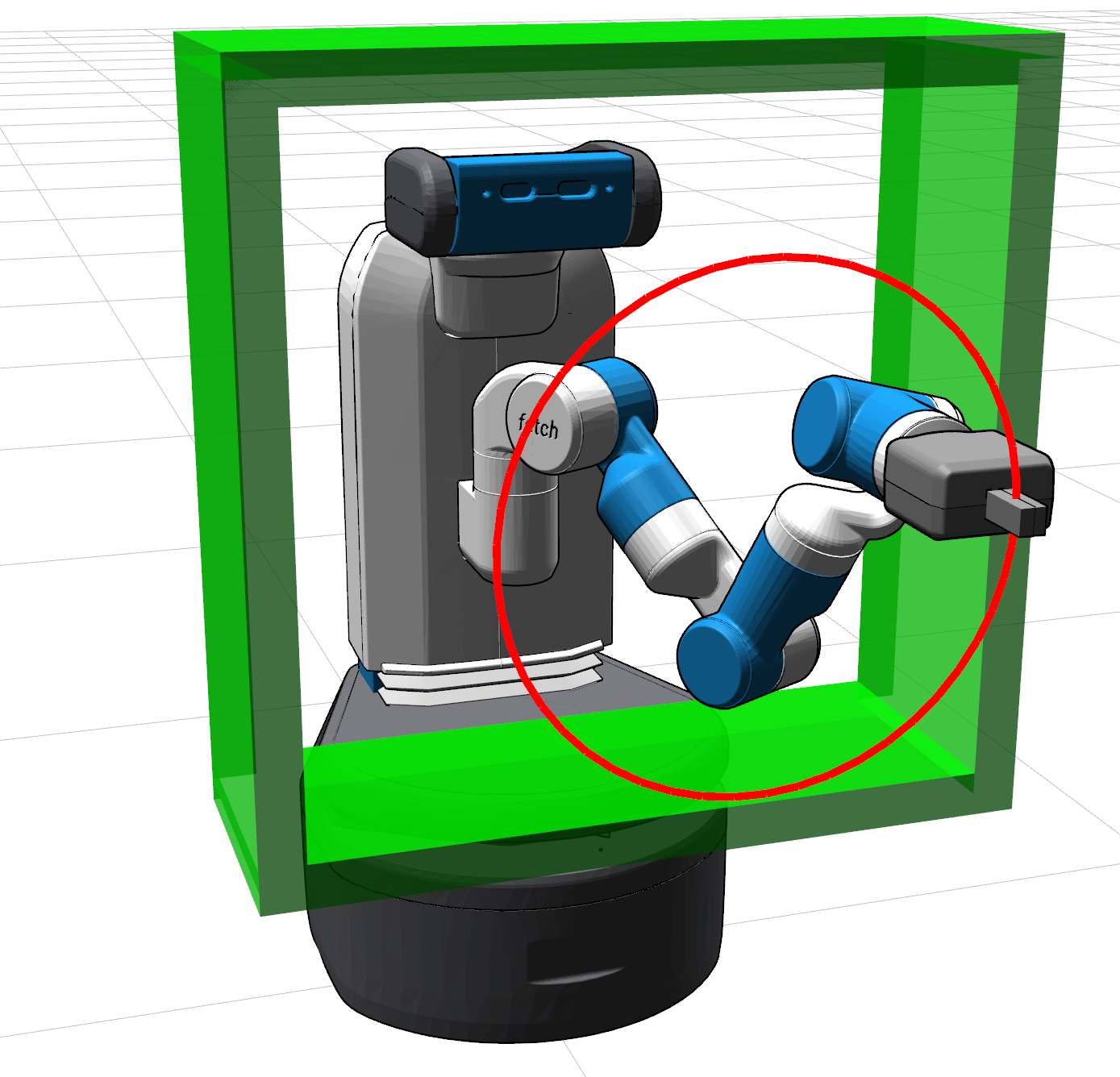}
        \caption{Fetch.Full - Circle}
    \end{subfigure}
    \hfill
    \begin{subfigure}[b]{0.23\textwidth}
        \centering
        \includegraphics[width=\textwidth]{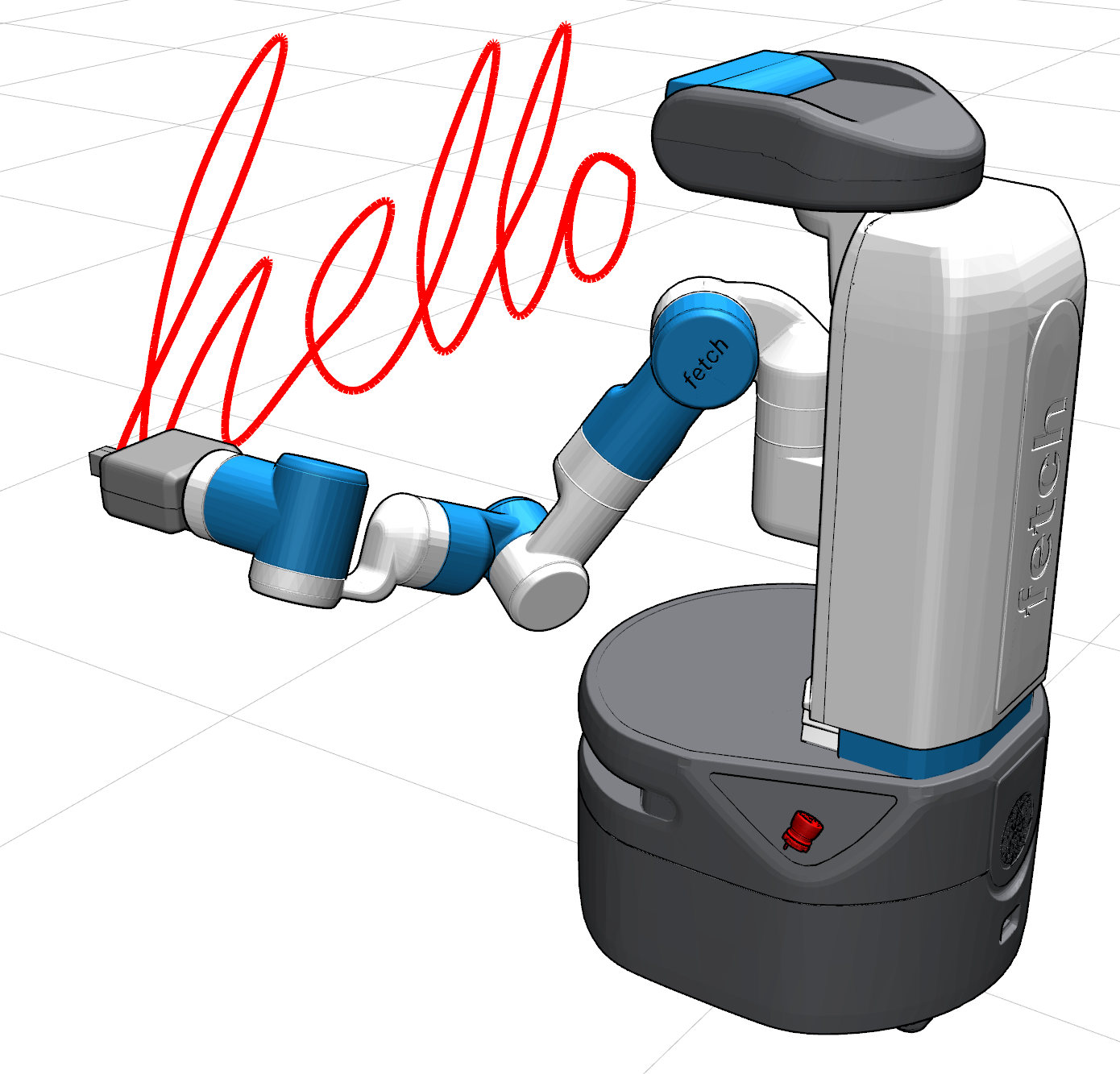}
        \caption{Fetch.Full - Hello}
    \end{subfigure}
    \hfill
    \begin{subfigure}[b]{0.23\textwidth}
        \centering
        \includegraphics[width=\textwidth]
        {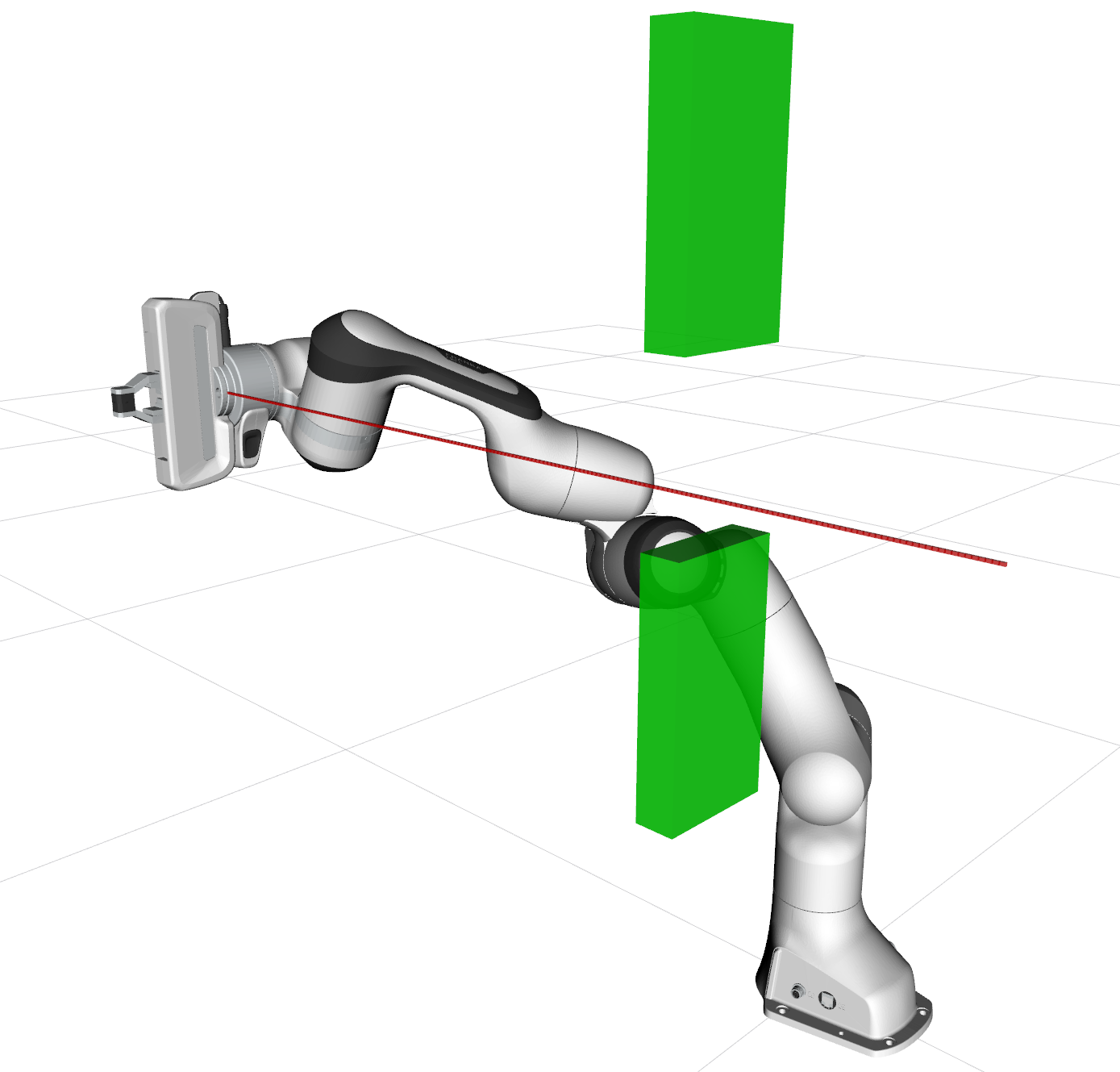}
        \caption{Panda - flappy-bird}
    \end{subfigure}
    \hfill
    \begin{subfigure}[b]{0.23\textwidth}
        \centering
        \includegraphics[width=\textwidth]{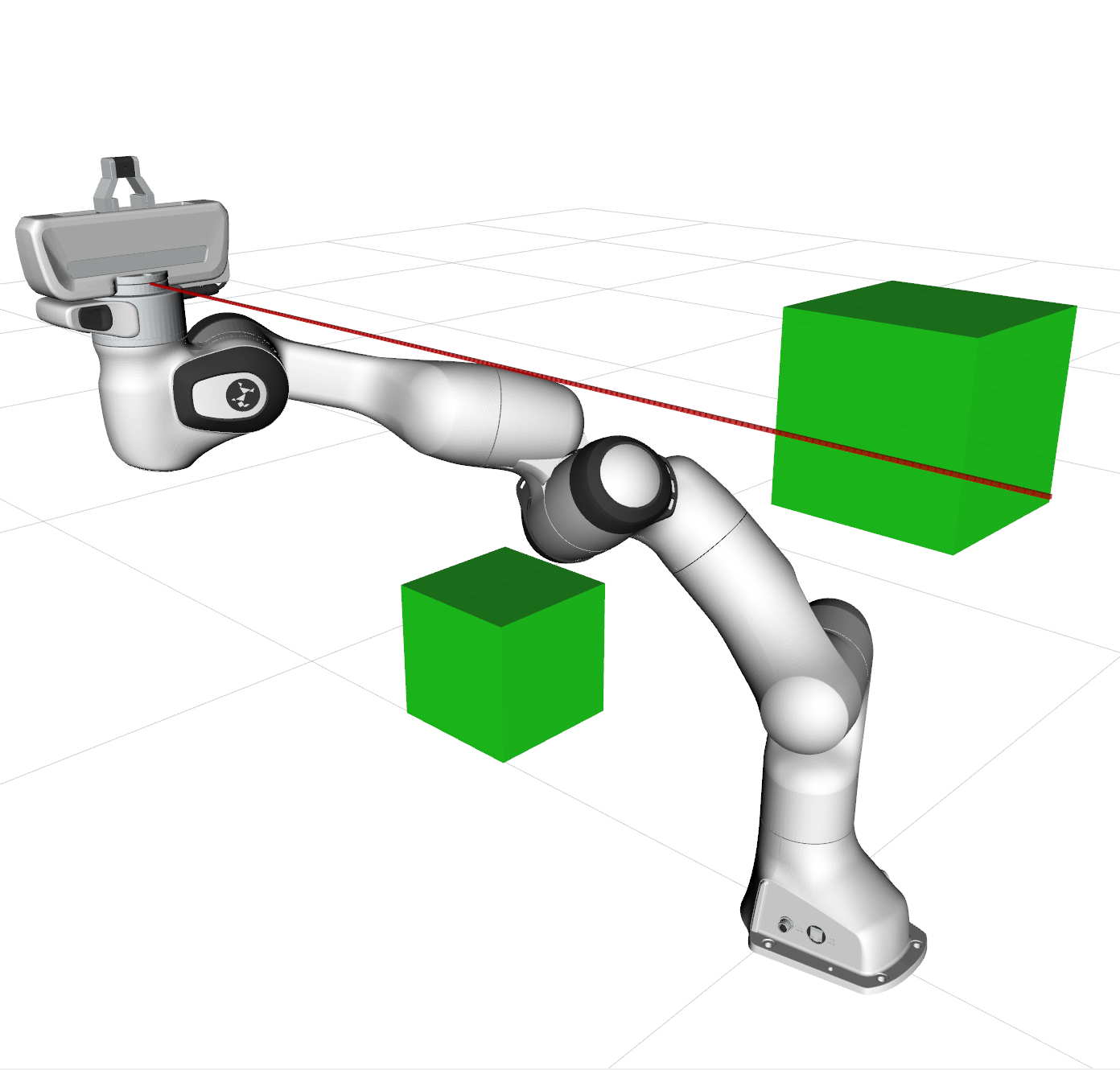}
        \caption{Panda - 2cubes}
    \end{subfigure}
    \vspace{5pt}
       \caption{Visualizations of a subset of the problems evaluated on.}
        \label{fig:problem-vizs}
\end{figure*}

% \begin{wrapfigure}{r}{0.4\textwidth}
\begin{figure}[t]
    \centering
    \includegraphics[width=0.3\textwidth]{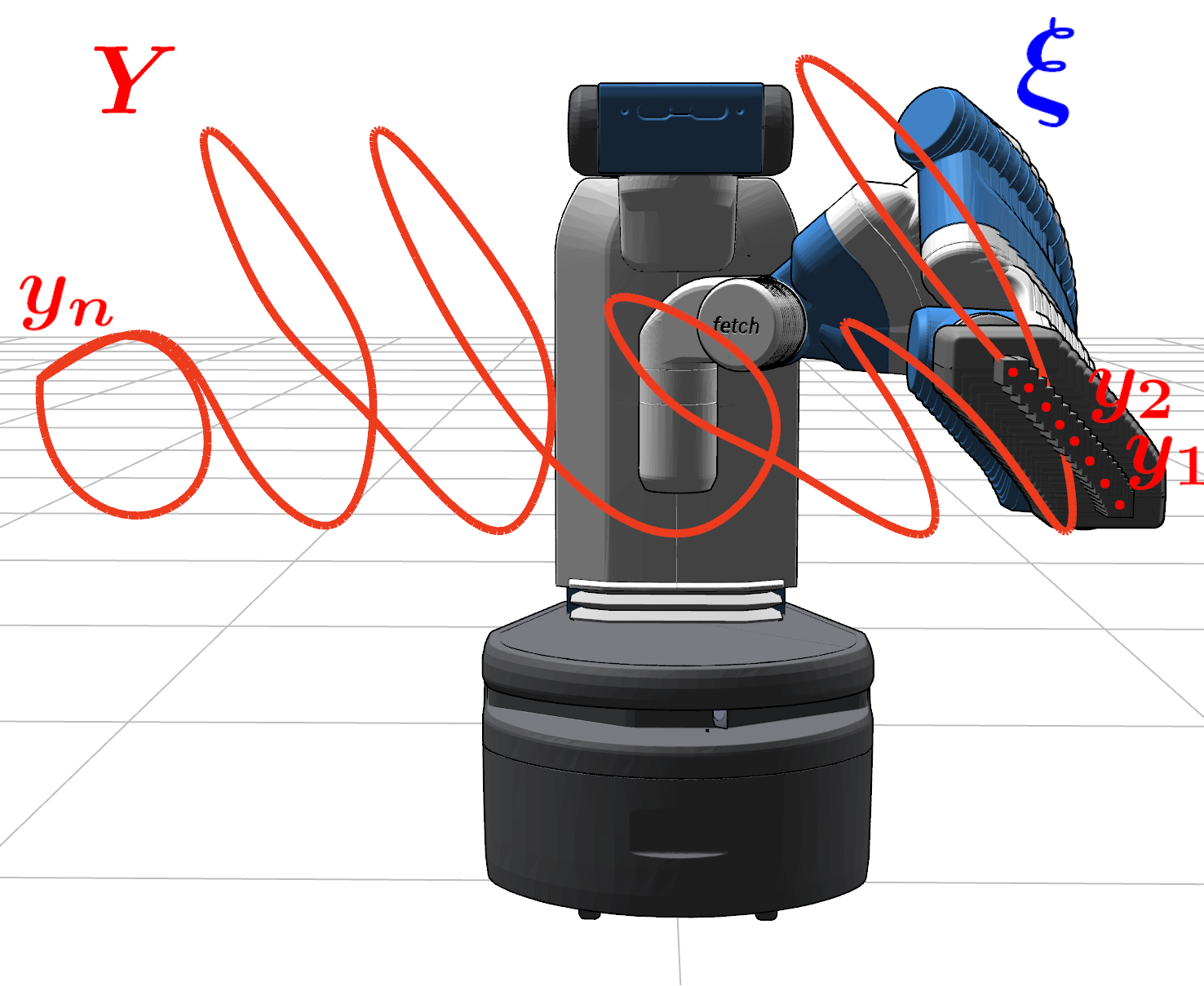}
    \caption{The `Fetch.Arm - hello' problem visualized. The sweeping motion of the robots arm is generated by overlaying the first 20 configurations from a trajectory $\xi$ returned by CppFlow. The red cursive `hello' is the target path, denoted $\boldsymbol{Y}$. The target path $\boldsymbol{Y}$ is composed of $n$ target poses $y_1, ..., y_n$}
    \label{fig:problem_specification}
% \end{wrapfigure}
\end{figure}

\textbf{Pose error}. At every timestep $i$, the deviation of the positional and rotational components of the robots end effector from the target pose must be within the mechanical repeatability of the robot. Specifically, positional error is the euclidean norm of the positional difference, and rotational error is the geodesic distance between the rotational components of $\text{FK}(q_i)$ and $y_i$. A positional error threshold of 0.1 mm is assigned, as this is the reported positional mechanical repeatability of both the Franka Panda and the Fetch Manipulator. The rotational error threshold is set to 0.1 deg, which is found through a procedure that finds the expected maximum rotational error for a configuration within the region of positional repeatability. Here and in the rest of the paper, the subtraction between poses in $\textrm{SE}(3)$ indicates elementwise subtraction between the x, y, z, and roll, pitch, and yaw components of each pose.

% found by calculating $\text{FK}(q_i) - y_i$, 

% must be within the mechanical repeatability of the robot. Here and in the rest of the paper, the subtraction between poses in $\textrm{SE}(3)$ indicates elementwise subtraction between the x, y, z, and roll, pitch, and yaw components of each pose. 

% To find the rotational error threshold, a procedure was performed that estimates the rotational repeatability of a robot as a function of the positional repeatability and the robots kinematics. A value of 0.1 deg was found through this procedure, which is used the threshold value.

% The approach is detailed in the appendix. 
% The rotational error threshold is set to 0.1 deg, which is found through a procedure that finds the expected maximum rotational error for a configuration within the region of positional repeatability. This procedure is presented in the Appendix.

% For example, collisions between child links and their parents are not checked, given that they won't collide when their connecting joint is within its limits. 
\textbf{Collision avoidance}. At no timestep in the trajectory can the robot collide with itself or the environment. A virtual capsule is affixed to each link of the given robot \textit{a priori}. Each capsule is the minimum enclosing capsule for a given links visual mesh and is found through a Quadratic Programming optimization procedure. Given capsules for each link, self collision checking is performed by evaluating whether any two capsules are intersecting. Checks between certain links are removed when it is impossible for these links to collide if joint limits are respected. The exact position and shape of every obstacle in the environment is assumed to be provided. In a similar manner, robot-environment collisions are found by checking for intersections between any link's capsule and the obstacles in the environment. Obstacles are all cuboids in this work for simplicity. It is assumed that the time parameterization between configurations is small enough such that collision checks between timesteps are not required.

\textbf{Joint limits}. The configuration $q$ at each timestep $i$ must be within the robots joint limits.
% Note that certain joints on the Fetch manipulator are continuous rather than revolute, meaning that their range is $(-\inf, \inf)$ however in practice we limit their range to $[-\pi, \pi]$ for simplicity

\textbf{Discontinuities}. The configurations across two different timesteps must stay close to one another as large changes in configuration space may not be achievable by the robot. A value of 7 degrees and 2 cm is chosen as the maximum absolute distance an individual revolute or prismatic joint may change across a timestep, respectively.

Lastly, while not a constraint, \textit{trajectory length} is an important metric for evaluating the quality of a trajectory. Trajectory length is defined as the cumulative change in configuration-space across a trajectory (measured in radians and meters) and gives an estimate for how long the trajectory will take to execute on an actual robot. The \textit{time to find a valid solution} is an important metric as well. As the name suggests this is the total elapsed time before a constraint satisfying trajectory is found.% This has implications in an industrial setting for example, where longer trajectories will reduce the lifetime of robots.

% Convergence
\begin{figure*}[t]
    \centering
    \vspace{0.1cm}
    \input{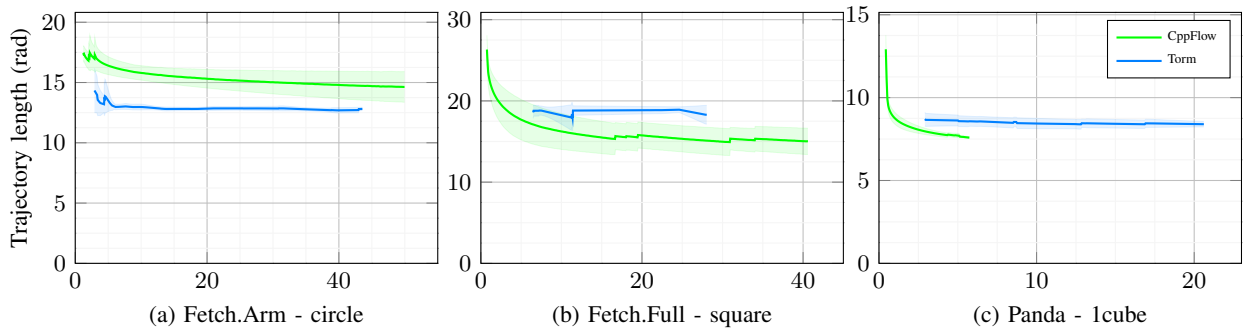}
    \caption{Trajectory length (rad) convergence results for CppFlow and Torm. Stampede results are ommited as the planner is not anytime - it returns only a single trajectory. A lower value is better, as this indicates a trajectory that can be executed in a shorter time. Plots are generated by averaging the trajectory length convergence curves of the designated planner from 10 planning runs. The mean time to a valid solution time is the first time instant plotted. The convergence behavior for CppFlow is strictly better for `Fetch.Full - square' and `Panda - 1cube', as measured by initial valid solution time and asymptotic limit. CppFlow has a 1.169x larger final trajectory length for `Fetch.Arm - circle' problem as compared to Torm, however, it takes Torm 1.83x as long for Torm to find an initial valid trajectory for this problem. The results indicate CppFlow and TORM perform comparably on this metric. CppFlow performs favorably in terms of `Time fo valid solution' and `Initial solution time' however.}

    \vspace{-0.5\baselineskip}

    % \textcolor{blue}{The results indicate CppFlow and TORM perform comparably on this metric. CppFlow performs favorably in terms of `Time fo valid solution' and `Initial solution time' however.}}

    \label{fig:convergence_plot}

    % \label[fig]{fig:convergence_plot}

    % motivation for using mean start/end times: in this plot we are looking at successful runs only. failed runs have no convergence results so aren't included. because of this we can use mean as every run will have a t0 and tf we can use.

    % Final TL (rad)
    % fetch.Arm - circle: 14.98 vs 12.81  -> 'cppflow has a 16.9399% increase in TL'
    % fetch.Arm - circle: 14.98 vs 12.81  -> 'cppflow has a 1.169398907x larger TL'

    % TVS (MEAN)
    % fetch.Arm - s:       vs  -> Torm/cpf:  
    % fetch.Arm - square:  vs   -> Torm/cpf: 
    % fetch.Arm - circle: 1.6084465503692627 vs 2.9553559  -> Torm/cpf: 1.83
\end{figure*}

\section{CppFlow}

We now present the CppFlow method. It is composed of three main components: a candidate motion plan generator, a global discrete search procedure, and a trajectory optimizer.

\subsection{Candidate Motion Plan Generation}

CppFlow begins by generating $K$ approximate motion plans using IKFlow, a learned generative IK solver ($K$ is a hyperparameter set here to 175)~\cite{ames2022ikflow}. IKFlow models are trained by fitting a Conditional Normalizing Flow model to a dataset of ground truth IK solutions. To bias the model away from self collisions, the IK solutions in the dataset are all non self colliding. IKFlow does not consider obstacles in the environment. The training procedure is covered in detail in prior work~\cite{ames2022ikflow, Papamakarios2019NormalizingInference}. IKFlow generates IK solutions as a function of a latent vector $z$ and a target cartesian pose $y$, in batch on the GPU. IKFlow models empirically exhibit a desirable property: small changes to either the latent vector or target pose result in small changes to returned IK solutions. We use IKFlow as the generative-IK model because of this smoothness property and its low inference time and high accuracy.

% Owing to the diversity of the dataset and the performance of the underlying generative model, returned solutions approximately cover the entire solution space.
% \vspace{-0.5\baselineskip}

To generate the $K$ motion plans, $K$ latent vectors are drawn uniformly at random from a unit hypercube, and repeated $n$ times. Randomizing the latent vectors ensures a diverse set of IK solutions and resulting motion plans are returned. Each set of $n$ repeated latent codes is paired with a copy of the target path $\boldsymbol{Y}$. IKFlow is then called with this batch of $nK$ latent vector - pose pairs. Lastly, the Candidate motion plans are generated by segmenting the returned IK solutions according to their corresponding latent vector. Advantageously, the returned IK solutions for each of the $K$ approximate motion plans are found to change slowly. This is because the latent vector is held fixed and the target path is slowly changing, satisfying the conditions for the smoothness property stated above. The runtime of this procedure scales linearly with the hyperparameter $K$, target path length $n$, and as a function of the performance characteristics of the GPU.

\newcolumntype{F}[1]{%
    >{\raggedright\arraybackslash\hspace{0pt}}p{#1}}%
\newcolumntype{T}[1]{%
    >{\centering\arraybackslash\hspace{0pt}}p{#1}}%

% llF{0.1\textwidth}F{0.1\textwidth}F{0.1\textwidth}F{0.1\textwidth}F{0.1\textwidth}

\begin{table*}[t]
    \centering
    \renewcommand{\arraystretch}{1.05}

    \vspace{0.1cm}
\begin{tabular}{lF{0.1\textwidth}F{0.09\textwidth}F{0.06\textwidth}F{0.081\textwidth}F{0.1\textwidth}F{0.081\textwidth}F{0.1\textwidth}}
%\toprule
\textbf{Problem} & \textbf{Planner} & \textbf{Time to valid solution (s)} & \textbf{Initial solution time (s)} & \textbf{Planning success rate (\%) - 2.5s max} & \textbf{Trajectory length (rad/m) - 2.5s max} & \textbf{Planning success rate (\%) - 50s max} & \textbf{Trajectory length (rad/m) - 50s max} \\
\midrule

\multirow[t]{3}{9em}{Fetch.Arm - circle} & $\textrm{cppflow - ours}$ & \textbf{1.211} & \textbf{0.014} & \textbf{90.0} & 17.39 & \textbf{100.0} & 14.98 \\
  & $\textrm{stampede}^*$ & 91.566 & 91.566 & 0.0 & --- & \textbf{100.0} & \textbf{11.105} \\
  & torm & 2.916 & 0.715 & 50.0 & \textbf{14.82} & \textbf{100.0} & 12.81 \\
\hline
\multirow[t]{3}{9em}{Fetch.Arm - hello} & $\textrm{cppflow - ours}$ & \textbf{1.126} & \textbf{0.016} & \textbf{100.0} & 64.46 & \textbf{100.0} & 62.52 \\
  & $\textrm{stampede}^*$ & $\inf$ & $\inf$ & 0.0 & --- & 0.0 & --- \\
  & torm & 3.598 & 1.414 & 30.0 & \textbf{62.426} & \textbf{100.0} & \textbf{58.653} \\
\hline
\multirow[t]{3}{9em}{Fetch.Arm - rotation} & $\textrm{cppflow - ours}$ & \textbf{0.535} & \textbf{0.014} & \textbf{100.0} & \textbf{30.423} & \textbf{100.0} & \textbf{26.758} \\
  & $\textrm{stampede}^*$ & $\inf$ & $\inf$ & 0.0 & --- & 0.0 & --- \\
  & torm & 1.27 & 1.066 & 70.0 & 30.5 & \textbf{100.0} & 27.13 \\
\hline
\multirow[t]{3}{9em}{Fetch.Arm - s} & $\textrm{cppflow - ours}$ & \textbf{0.788} & \textbf{0.014} & \textbf{100.0} & 17.18 & \textbf{100.0} & 15.47 \\
  & $\textrm{stampede}^*$ & 137.569 & 137.569 & 0.0 & --- & \textbf{100.0} & \textbf{10.856} \\
  & torm & 1.024 & 0.844 & 80.0 & \textbf{13.549} & \textbf{100.0} & 12.77 \\
\hline
\multirow[t]{3}{9em}{Fetch.Arm - square} & $\textrm{cppflow - ours}$ & \textbf{0.747} & \textbf{0.014} & \textbf{100.0} & 21.71 & \textbf{100.0} & 18.48 \\
  & $\textrm{stampede}^*$ & 110.774 & 110.774 & 0.0 & --- & \textbf{100.0} & \textbf{14.841} \\
  & torm & 1.045 & 0.79 & \textbf{100.0} & \textbf{17.287} & \textbf{100.0} & 16.56 \\
\hline
\multirow[t]{2}{9em}{Fetch.Full - circle} & $\textrm{cppflow - ours}$ & \textbf{1.02} & \textbf{0.011} & \textbf{100.0} & \textbf{20.13} / \textbf{0.37} & \textbf{100.0} & \textbf{13.28} / \textbf{0.46} \\
  & torm & 14.595 & 2.363 & 10.0 & 21.99 / 0.46 & 90.0 & 20.67 / 0.45 \\
\hline
\multirow[t]{2}{9em}{Fetch.Full - hello} & $\textrm{cppflow - ours}$ & \textbf{1.243} & \textbf{0.011} & \textbf{70.0} & \textbf{49.74} / \textbf{2.9} & \textbf{100.0} & \textbf{39.97} / \textbf{3.49} \\
  & torm & $\inf$ & 4.12 & 0.0 & --- & 20.0 & 76.07 / 2.39 \\
\hline
\multirow[t]{2}{9em}{Fetch.Full - rotation} & $\textrm{cppflow - ours}$ & \textbf{0.605} & \textbf{0.011} & \textbf{100.0} & \textbf{28.47} / \textbf{1.01} & \textbf{100.0} & \textbf{21.8} / \textbf{1.24} \\
  & torm & 15.31 & 1.784 & 0.0 & --- & \textbf{100.0} & 29.15 / 0.77 \\
\hline
\multirow[t]{2}{9em}{Fetch.Full - s} & $\textrm{cppflow - ours}$ & \textbf{0.785} & \textbf{0.011} & \textbf{100.0} & \textbf{22.08} / \textbf{0.55} & \textbf{100.0} & \textbf{14.14} / \textbf{0.77} \\
  & torm & $\inf$ & $\inf$ & 0.0 & --- & 0.0 & --- \\
\hline
\multirow[t]{2}{9em}{Fetch.Full - square} & $\textrm{cppflow - ours}$ & \textbf{0.751} & \textbf{0.011} & \textbf{100.0} & \textbf{19.65} / \textbf{0.35} & \textbf{100.0} & \textbf{13.97} / \textbf{0.39} \\
  & torm & 9.255 & 2.997 & 0.0 & --- & 70.0 & 18.9 / 0.6 \\
\hline
\multirow[t]{2}{9em}{Panda - 1cube} & $\textrm{cppflow - ours}$ & \textbf{0.434} & \textbf{0.011} & \textbf{100.0} & \textbf{8.075} & \textbf{100.0} & \textbf{7.585} \\
  & torm & 2.912 & 2.585 & 0.0 & --- & \textbf{100.0} & 8.13 \\
\hline
\multirow[t]{2}{9em}{Panda - 2cubes} & $\textrm{cppflow - ours}$ & \textbf{0.484} & \textbf{0.011} & \textbf{100.0} & \textbf{9.627} & \textbf{100.0} & \textbf{9.031} \\
  & torm & 62.478 & 39.346 & 0.0 & --- & 40.0 & 13.21 \\
\hline
\multirow[t]{2}{9em}{Panda - flappy-bird} & $\textrm{cppflow - ours}$ & \textbf{0.63} & \textbf{0.011} & \textbf{100.0} & \textbf{20.323} & \textbf{100.0} & \textbf{20.323} \\
  & torm & $\inf$ & $\inf$ & 0.0 & --- & 0.0 & --- \\
\bottomrule
\end{tabular}

    % {\raggedright Median \par}

    \caption{Results from CppFlow, Torm, and Stampede on the 13 test problems. `Time to Valid Solution' - median time to find a trajectory with less than 1mm/.1deg pose error and all other constraints satisfied. `Initial Solution Time' - median time to find a trajectory, regardless of its validity. Failed runs count for $\inf$ planning time for both `Time to Valid Solution' and `Initial Solution Time'. `Planning Success Rate, \textit{x} seconds' - the percentage of runs which have found a valid trajectory before \textit{x} seconds have elapsed. `Trajectory length (rad/m), \textit{x} seconds': The mean cumulative change in configuration-space of the revolute, prismatic joints of the latest valid trajectory found before \textit{x} seconds have elapsed. A lower trajectory is better, as generally this means the robot can execute the trajectory faster. The astricks on the $\textrm{stampede}^*$ rows indicates that the planning success rate and trajectory length is for the final returned trajectory, which may be returned after 50s. Stampede failed to run at all with the Fetch.Full kinematic chain so was excluded from the results. CppFlow returns an initial solution faster than either of the other methods on all problems. It will take Torm and Stampede 19.84x and 132.8x as long on average respectively to find a valid solution.}

    \vspace{-0.75\baselineskip}

    \label{table:planning_time} % Note: Label needs to come (after?) before caption!!

\end{table*}

% \vspace{-0.1\baselineskip}

\subsection{Global discrete search}
The objective of the global discrete search module is to find the optimal $\xi$ as measured by collision avoidance and the \textit{Maximum Joint Angle Change} ($\texttt{mjac}$) - the maximum absolute change in joint angle between any two timesteps along a trajectory. Formally, $\texttt{mjac}=\max(\max(|q_{i+1}[j] - q_{i}[j]| \forall j \in [1, ..., d]) \forall i \in [1, ..., n])$.

The module first creates a directed graph from the $K$ motion plans. Each IK solution is considered a node. Edges are connected from each IK solution to the $K$ IK solutions at the following time step. A dynamic-programming-based search procedure is run on the graph. The cost at each node is set to the minimum $\texttt{mjac}$ achievable if the node is included in the returned path. To avoid trajectories that are close to the joint limits, an additional cost of 10 (which is greater than the largest possible \texttt{mjac} cost of $\pi$ for revolute joints, and the range of Fetch.Full's prismatic joint) is added to configurations that are within 1.5 degrees/3 cm of their joint limits. An additional cost of 100 is added to configurations that would lead to a collision. IKFlow does not perform any collision checking itself, so a separate collision checking module checks for these collisions. Upon completion, a backtracking procedure is executed, which returns $\xi_{\textrm{search}}$. 

% \textcolor{red}{this sentence is good but fluff:} The returned $\xi_{\textrm{search}}$ is optimally smooth (as measured by $\texttt{mjac}$), free of obstacles, and far from joint limits which makes it an ideal seed for a trajectory optimizer. \textcolor{red}{A similar approach is taken in~\cite{rakita_stampede_2019}, however in this other work a --- is not performed}.

% the  which is optimal according to this cost specification

Trajectories with large $\texttt{mjac}$ values are likely to fail in the next step, as this indicates the existence of large joint space discontinuities, which are difficult to optimize. To remedy this, after the search finishes, the $\texttt{mjac}$ of $\xi_{\textrm{search}}$ is calculated. If it is above a threshold (12 deg/3 cm), an additional set of motion plans is generated and added to the existing set before repeating the search. Collision and joint limit checks performed on the initial plans are retained to conserve computation. This is an effective approach to prevent bad optimization seeds, based on the intuition that a denser covering of the relevant portions of configuration space likely contains a better configuration space path.

% The search algorithm is efficiently implemented using GPU accelerated functions. In practice it takes around 0.5 second to search over a 500 timestep path with $K=175$. \textcolor{red}{can rm this line}

% The runtime of the search is in O($nK$) is efficiently implemented using GPU accelerated functions. In practice it takes around 0.5 second to search over a 500 timestep path with $K=175$.

% \vspace{-0.1\baselineskip}

% \subsection{Trajectory Optimization}
\subsection{Levenberg-Marquardt for Trajectory Optimization}

% The trajectory optimizer reduces the pose error of the end effector and the trajectory length while ensuring robot-robot collision, robot-environment collision, and joint limit constraints are satisfied.

The optimization problem is framed as a nonlinear least squares problem and solved using the Levenberg-Marquardt algorithm. As opposed to stochastic gradient descent and other first-order optimization methods previously used to solve this problem~\cite{kang2020torm}, Levenberg-Marquardt approximates the objective Hessian matrix with first derivative information in order to make significantly larger and more accurate steps on least-squares objectives. As a result, when initial seeds are in close vicinity to a valid solution, only 1 to 3 steps are required to find a valid solution.

The optimizer keeps track of the latest valid solution;  this can be returned if the user wants to exit making CppFlow an \textit{anytime} planner i.e., it continuously improves its plan until stopped. If a valid solution is not found, the dynamic programming search is repeated using additional motion plans from IKFlow. This lowers the $\texttt{mjac}$ of the $\xi$ used as a seed, increasing the chance it will be successfully optimized. 

% Only a single retry is performed before the run is counted as a failure. \textcolor{red}{j: consider removing this}

Similar to Torm, a non-stationary objective function is used, leading to improved convergence performance~\cite{kang2020torm}. The optimizer switches between optimizing only for pose error and for trajectory length and obstacle avoidance. To balance the frequency of the two, the optimizer optimizes for pose error exclusively until the positional and rotational error of the end effector is below the respective specified threshold at every timestep. A trajectory length and obstacle avoidance optimization step is then taken, after which the optimizer switches back to pose only, and the cycle continues. This ordering ensures that the trajectory stays close to a valid solution throughout the optimization process. The residual terms are $r_{\textrm{pose}}$ for pose error, and $r_\textrm{diff}$ for trajectory length and obstacle avoidance error. Figure~\ref{eq:lm_residual} shows the composition of each residual. The residual components are listed below.

% residual
\begin{figure}[t]
    \centering
    \setlength{\columnsep}{0cm}
    \begin{multicols}{2}
        \noindent
        \begin{equation}
            % r_{\textrm{pose}}(\xi) = \begin{bmatrix}
            r_{\textrm{pose}} = \begin{bmatrix}
                \textrm{FK}(q_1).x - y_1.x \\
                \textrm{FK}(q_1).y - y_1.y \\
                \textrm{FK}(q_1).z - y_1.z \\
                \textrm{FK}(q_1).\Phi - y_1.\Phi \\
                \textrm{FK}(q_1).\Theta - y_1.\Theta \\
                \textrm{FK}(q_1).\Psi - y_1.\Psi \\
                \dots \\
                % \textrm{FK}(q_n).x - y_n.x \\
                % \textrm{FK}(q_n).y - y_n.y \\
                % \textrm{FK}(q_n).z - y_n.z \\
                % \textrm{FK}(q_n).\Phi - y_n.\Phi \\
                % \textrm{FK}(q_n).\Theta - y_n.\Theta \\
                \textrm{FK}(q_n).\Psi - y_n.\Psi \\
            \end{bmatrix}\nonumber
        \end{equation}
        % \break
        \begin{equation}
            % r_\textrm{diff}(\xi) = \begin{bmatrix}
            r_\textrm{diff} = \begin{bmatrix}
                q_2 - q_1 \\
                q_3 - q_2 \\
                \dots \\
                q_n - q_{n-1} \\
                % \rule{1cm}{0.4pt} \\
                \textrm{self-coll-dist}(q_1) \\
                \dots \\
                \textrm{self-coll-dist}(q_n) \\
                % \rule{1cm}{0.4pt} \\
                \textrm{env-coll-dist}(q_1) \\
                \dots \\
                \textrm{env-coll-dist}(q_n) \\
            \end{bmatrix}\nonumber
        \end{equation}
    \end{multicols}
    % \vspace{-0.1\baselineskip}
    \caption{The residual term used by the optimizer. The symbols $\Phi$, $\Theta$, and $\Psi$ represent the roll, pitch, and yaw of the pose.}
    \label{eq:lm_residual} % Note: Label needs to come (after?) before caption!!
\end{figure}

\textbf{Pose error} The pose of the end effector at each timestep is calculated by a Forward Kinematics function using an efficient custom PyTorch implementation which performs calculations in parallel on the GPU. A residual term for the error in the x, y, z, and roll, pitch, and yaw dimensions ($\textrm{FK}(q_i).x - y_i.x$, ..., $\textrm{FK}(q_i).\Psi - y_i.\Psi$) are added to $r_{\textrm{pose}}$, for every $q_i$ in $\xi$. The Jacobian of these residuals wrt $\xi$ is found by observing that in the derivative of $\textrm{FK}(q_i).x - y_i.x$ wrt $q_i$, the constant term $y_i.x$ goes to 0 so the derivative is $J_{\textrm{fk}}(q_i).x$. A second custom PyTorch implementation quickly calculates this kinematic Jacobian term in parallel on the GPU. Once the residual and Jacobian are calculated, the Levenberg-Marquardt update $(J^T J + \lambda I) \Delta \xi = J r_{\textrm{pose}}$ is calculated using PyTorch's batched LU decomposition functionality.

% \vspace{-0.6\baselineskip}

% An observation can be made that the pose error at each timestep is independent from other timesteps, which gives rise to a block diagonal structure of the jacobian of $r_{\textrm{pose}}$. Using this structure, the Levenberg-Marquardt update $(J^T J + \lambda I) \Delta \xi = J r_{\textrm{pose}}$ is solved at each timestep independently.
\textbf{Trajectory length}. A differencing loss is used as a surrogate error for trajectory length. It penalizes changes in joint angle across consecutive timesteps: $q_{i+1} - q_i \forall i \in [1, ..., n-1]$, which encourages configurations to stay close to one another. The Jacobian of each differencing error $(q_{i+1} - q_i)$ wrt $\xi$ is the identity matrix for configuration $q_{i+1}$, and the negated identity matrix for $q_i$.

\vspace{-0.1\baselineskip}

\textbf{Robot-robot collision avoidance}. For self-collision checking, each robot is represented as a set of capsules with endpoints $(a, b)$ and radius $c$. As in \cite{safeea2018efficient}, we formulate the minimum distance between a capsule pair $(a_1, b_1, c_1)$ and $(a_2, b_2, c_2)$ as the minimum cost, less $c_1 + c_2$, of a convex quadratic program
\begin{align*}
    \min_{t_1, t_2} \quad &|| (a_1 + (b_1 - a_1) t_1) - (a_2 + (b_2 - a_2) t_2) ||_2^2 \\
    \text{subject to} \quad &0 \le t_1, t_2 \le 1
\end{align*}
which we solve with a custom batched active-set method implementation. While the solutions of convex programs are differentiable~\cite{amos2017optnet}, since we evaluate many collision pairs for the same joint configuration, it is efficient to compute Jacobians by batched forward-mode~\cite{siskindperturbation} automatic differentiation.
% yo, for some reason there's two lines of white space between these two

\textbf{Joint limits}. During optimization, joint limit constraints are enforced by clamping $\xi$ to the robot joint limits.

% It is also assumed that the time parameterization is fine enough such that collision checks between timesteps are not required.
\textbf{Robot-environment collision avoidance}. We represent obstacles as cuboids with arbitrary spatial positioning. We formulate the minimum distance between a capsule and a cuboid in a frame axis-aligned to the cuboid, in which the capsule has endpoints $(a, b)$ and radius $r$, and the cuboid has extents $p_{min} = (x_{min}, y_{min}, z_{min})$ and $p_{max} = (x_{max}, y_{max}, z_{max})$ as the minimum cost, less $c$, of the convex quadratic program.
\begin{align*}
    \min_{t, p} \quad &|| p - (a + (b - a) t) ||_2^2 \\
    \text{subject to} \quad &0 \le t \le 1, \textrm{  } p_{min} \le p \le p_{max} \\
    % \text{subject to} \quad &0 \le t \le 1 \\
    % & p_{min} \le p \le p_{max}
\end{align*}

\vspace{-1.8\baselineskip}

\section{Experiments \& Analysis}

CppFlow is evaluated on 13 different planning problems and 3 different robots. Other SOTA methods, including Torm and Stampede~\cite{kang2020torm,rakita_stampede_2019} are also evaluated to provide a comparison. The Fetch.Arm problems are the same as reported in  ~\cite{kang2020torm}. The Fetch.Full problems are the same except for the addition of the Fetch prismatic joint to the kinematic chain. The Panda problems were created in this work.

We grade the planner on three axes: success rate, time to get an initial valid solution, and trajectory length over time. Metrics include the time to find a valid solution (maximum of .1 mm position and .1 degree rotation error without violating any other constraints), initial solution time (regardless of validity), planning success rate, and trajectory length. For CppFlow and Torm, which are anytime, these are reported throughout the optimization whereas for Stampede only the final returned trajectory is analyzed.

\vspace{-0.15\baselineskip}

\subsection{Experiments}

There are four obstacle-free problems, including `hello', `rotation' for both Fetch.Full and Fetch.Arm. The problems with obstacles include `circle', `s', and `square` for Fetch.Full and Fetch.Arm and the three Panda problems. These are visualized in~\ref{fig:problem-vizs}. While Torm allows for an initial joint configuration to be set, this feature is disabled for a fair comparison. Further, the target path for the `rotation' problems provided in the Torm GitHub repository had 40 additional waypoints added to reduce the distance between the poses of the target path. The poses still parameterize the same motion, however. We run Stampede on the obstacle problems even though it does not account for obstacles as a further point of comparison - the obstacles are ignored in these cases. A software bug in the provided code prevented Stampede from running on the Fetch.Full robot.

% Further, the target path for the `rotation' problems provided in the Torm github repository was re-interpolated with 40 additional waypoints to reduce the distance between the poses of the target path. Stampede is still run on the obstacle problems even though the planner does not account for obstacles as a further point of comparison - the obstacles are simply ignored in these cases. Further, a software bug in the provided code prevented Stampede from running on the Fetch.Full robot at all.

% The poses still parameterize the same motion however. 

% Further, the target path for the `rotation' problems provided in the Torm github repository had 40 additional waypoints added to reduce the distance between the poses of the target path. The poses still parameterize the same motion however. Stampede is still run on the obstacle problems even though the planner does not account for obstacles as a further point of comparison - the obstacles are simply ignored in these cases. Further, a software bug in the provided code prevented Stampede from running on the Fetch.Full robot at all.

The three test robots are `FetchArm', `FetchFull', and `Panda' (7, 8, and 7 DOF, respectively). The difference between `FetchArm' and `FetchFull' is that `FetchFull' includes a prismatic joint (the `torso\_lift\_joint') which lifts the entire arm. Each planner is run 10X on each problem. CppFlow and Torm are allotted 60s per run, whereas Stampede has no time limit (it stops on its own after returning a solution). Tests are run on an Intel i9  with 20 CPUs, 124 GBs of RAM, and an Nvidia RTX 4090 graphics card.

% CppFlow and Torm both save the relevant metrics for all new trajectories. The time required to generate the metrics and save the results is ignored.

% For a fair comparison, the time required to evaluate the current best trajectory from Torm is subtracted from the planning time. This is not required for STAMPEDE because Stampede returns a single trajectory which can be analyzed after planning finishes. 
% Stampede does not account for external obstacles and therefore is omitted from the paths that include obstacles. Further, a bug was found in the implementation which causes the prismatic joint of the FetchFull manipulator to be incorrectly set, which offsets the end effector by a fixed amount from the target path. Therefore the results for STAMPEDE are omitted for FetchFull as well. 
% The exact specification of these robots will be made available upon acceptance to preserve the authors anonymity.

\vspace{-0.15\baselineskip}

\subsection{Results}

The numerical results are presented in Table~\ref{table:planning_time}; selected trajectory length convergence plots are shown in Figure~\ref{fig:convergence_plot}. 

CppFlow dominates on the first axis, planning success rate. The success rate is 100\% for all problems after 50s and reaches 96.7\% on average after only 2.5 seconds. In comparison, Torm fails to generate any plans more than 50\% of the time on three of the problems, and Stampede fails to generate any plans for 2/5 of its problems. This indicates CppFlow is a robust planner that is likely to succeed on difficult problems. CppFlow also dominates on the second axis: the time to get a valid solution. Valid solutions are generally found within 1 second and often in under 600ms. Compared to CppFlow, Torm takes between 1.29x to 129.15x as long to find its first valid solution. On average, it will take Torm 19.84x as long to find a valid solution. On the final axis, the trajectory length convergence behavior for CppFlow is strictly better for the Fetch.Full and Panda problems compared to Torm. The convergence results are roughly tied for 2/5 Fetch.Arm problems, while Torm has a lower asymptotic limit for the other 3 (including `Fetch.Arm - circle', which is shown in Figure~\ref{fig:convergence_plot}).
% , owing to the good initial guesses provided by IKFlow and the fast convergence of the Levenberg-Marquardt optimizer. 
The results indicate that CppFlow generates valid solutions faster than all other SOTA methods while producing trajectories of overall similar quality when run as an anytime planner. Crucially, CppFlow succeeds on the hardest problems, which indicates it is a highly capable planner. Given this, CppFlow would be a good all-around choice for a CPP planner, especially in settings that require quick planning times for unseen and potentially difficult planning problems, such as in a home or hospital. 

% \textcolor{red}{TODO(@jm) consider moving this final paragraph to the conclusion}

% is a capable and performant planner compared to other SOTA methods.

% \subsection{Limitations}
% \textcolor{red}{j: do we need this anymore?}
% Our system as described has several limitations.
% \textbf{Jerk Limiting}\quad CppFlow currently only minimizes the velocity of the final path, while many industrial robot controllers provide hard limits on joint-space jerk to preserve motors and gearboxes.
% \textbf{Realtime applicability}\quad CppFlow relies on nonconvex optimization for the the final solution refinement. Nonconvex optimization has few theoretical guarantees of convergence, much less in a fixed amount of time, making CppFlow inapplicable for planning scenarios with hard-realtime requirements.
% \textbf{Embeddability}\quad IKFlow provides a powerful system for generating initial solution guesses for pathwise-IK problems. However, it currently relies on large neural networks to generate these solutions. Such networks require GPU or other accelerator hardware that may not be present onboard robot hardware or in other compute-constrained environments.
% One limitation is that CppFlow assumes a different system will temporally parameterize the returned trajectories, and thus does specifically account for velocity, acceleration, or jerk limits.  

\vspace{-0.05\baselineskip}

\section{Related work}
% There exists work on a number of related topics to the CppFlow method presented, such as pathwise-IK, Generative modeling for inverse kinematics, is summarized.
% In this section the related work on topics relevant to the CppFlow method is summarized, which include pathwise-ik, generative modeling for IK, and Levenberg-Marquardt optimization for robotics.

% There have been a number of methods proposed to solve the pathwise-ik problem. 
% \subsection{Cartesian Path Planning}
% The Cartesian Path Planning problem has a rich history of proposed solutions. Search and optimization have been the primary paradigms, however recently learning methods have shown promise for improving the runtime and capability of systems. 

Torm solves CPP using gradient descent-based optimization to reach an acceptable solution~\cite{kang2020torm}. It  contains a candidate trajectory generator module that performs traditional IK along the reference trajectory to find a good approximate trajectory to optimize. RelaxedIK solves the problem of generating motion in real-time, which satisfies end-effector pose goal matching and motion feasibility~\cite{rakita_relaxedik_2018}. Stampede~\cite{rakita_stampede_2019} solves time-constrained CPP  by constructing and searching through a solution graph of viable configurations to find a satisfying trajectory. Luo and Hauser solve time-constrained CPP~\cite{6225341} through a decoupled optimization procedure in which pose error is first reduced before temporal matching is decoupled; however, only the position of the end effector is considered - orientation is ignored. While IK for redundant kinematic chains has been traditionally solved by using the Jacobian to iteratively optimize joint configuration, recent work has shown that generative modeling techniques can learn this mapping instead. These methods generate solutions in parallel on the GPU, enabling significantly better runtime scaling characteristics, albeit at the cost of accuracy. IKFlow~\cite{ames2022ikflow}, the solver used here, uses conditional Normalizing Flows to represent this mapping. The method in~\cite{9561589} learns the density of IK solutions at a given pose using a Block Neural Autoregressive Flow. NodeIK learns this mapping using a Neural Ordinary Difference Equation model~\cite{park_node_2022}. In~\cite{Ren2020LearningIK} Generative Adversarial Networks for IK are trained for the Fetch Manipulator and used for generating trajectories along a path; however, statistics on the pose error of returned solutions are not included. In~\cite{limoyo2022network}, the IK problem is reformulated as a distance geometry problem whose solutions are learned via Graph Neural Networks. A single GNN can produce IK solutions for multiple kinematic chains. However, the resulting solution accuracy for individual robots is lower than that of IKFlow and NodeIK.

\vspace{-0.05\baselineskip}
\section{Conclusion}
\vspace{-0.05\baselineskip}
We propose an efficient and capable anytime Cartesian Path Planner that combines techniques from search, optimization and learning to achieve SOTA results. Our planner achieves a 100\% success rate on a representative problem set and finds valid trajectories within 1.3 seconds, in the worst case. Additionally, it continually decreases trajectory length when time is allowed. Our results demonstrate the usefulness of generative IK models for kinematic planning and raise the question of where else they can be applied. We are exploring how CppFlow may be adapted for kinodynamic planning, which requires precise temporal specifications.

\hypersetup{breaklinks=true, colorlinks=false}

%===============================================================================
\newpage
\bibliographystyle{IEEEtran}

\bibliography{references}  % .bib

\end{document}